# Human Face Recognition from Part of a Facial Image based on Image Stitching

Osama R. Shahin, Rami Ayedi, Alanazi Rayan, Rasha M. Abd El-Aziz, Ahmed I. Taloba
Department of Computer Science, College of Science and Arts in Qurayyat, Jouf University, Saudi Arabia

*Abstract*—Most of the current techniques for face recognition require the presence of a full face of the person to be recognized, and this situation is difficult to achieve in practice, the required person may appear with a part of his face, which requires prediction of the part that did not appear. Most of the current forecasting processes are done by what is known as image interpolation, which does not give reliable results, especially if the missing part is large. In this work, we adopted the process of stitching the face by completing the missing part with the flipping of the part shown in the picture, depending on the fact that the human face is characterized by symmetry in most cases. To create a complete model, two facial recognition methods were used to prove the efficiency of the algorithm. The selected face recognition algorithms that are applied here are Eigenfaces and geometrical methods. Image stitching is the process during which distinctive photographic images are combined to make a complete scene or a high-resolution image. Several images are integrated to form a wide-angle panoramic image. The quality of the image stitching is determined by calculating the similarity among the stitched image and original images and by the presence of the seam lines through the stitched images. The Eigenfaces approach utilizes PCA calculation to reduce the feature vector dimensions. It provides an effective approach for discovering the lower-dimensional space. In addition, to enable the proposed algorithm to recognize the face, it also ensures a fast and effective way of classifying faces. The phase of feature extraction is followed by the classifier phase. Displacement classifiers using square Euclidean and City-Block distances are used. The test results demonstrate that the proposed algorithm gave a recognition rate of around 95%, to validate the proposed algorithm; it compared to the existing CNN and Multibatch estimator method.

## I. INTRODUCTION

Image stitching is the method used for consolidating various photographic images with an overlapping manner of view to get a sectioned display or high-resolution image. Most regular methodologies of image stitching require correct covers amongst images and indistinguishable exposures to create consistent results. Moreover, by using image stitching in computer vision and PC design applications, some digital cameras can stitch their photographs together internally [1]. Arrangement of the images may comprise at least two digital images taken of a solitary scene in various circumstances, from various sensors, or various perspectives. Image stitching strategies are classified into two general methodologies: feature-based techniques and direct techniques [2]. Feature-based techniques expect to identify a connection between the images through unmistakable features separated from the prepared images whereas direct techniques were dealing with all pixels of the parts of the image stitched. The feature-based strategy has the advantage of being more robust against direct techniques, and it can naturally find interlinkages between disorganized arrangements of images. Image stitching comprises three stages: Image Acquisition, Image Registration, and Image Blending.

Face recognition has wide applications in security, validation, surveillance, and distinct forensic evidence. Regular identification strategies such as ID cards and passwords are not considered as reliable as previously thought due to the various methods of hacking secret keys and others. As an option, biometrics, which is characterized by being a physical identification mark or belonging to a specific person, is not the same as others. The distinction between different individuals in the Known person database is the focal point of face recognition. In the future, face recognition is expected to oversee unlimited frameworks, such as access control for aircraft station security, smart home applications, structures, and faces, as well as for checking and monitoring buildings and vehicles and intelligent human-PC collaboration. In recent years, confrontational recognition has created limitless application fields from acknowledged assertion and proof to collaboration and correspondence between human devices and computers through video applications based on face recognition. Face recognition algorithms are regularly classified into three categories, which are comprehensive methods, feature-based methods, and hybrid methods.

The holistic methods category symbolizes the entire facial region as a high-dimensional vector that contributes to a classifier. Principle Component Analysis (PCA) is a successful agent technique for all-encompassing face recognition strategies, including Linear Discriminant Analysis (LDA) and Independent Component Analysis (ICA). Local methods, on the other hand, extract neighborhood characteristics from facial territories; for example, the eyes, mouth, nose, and cheeks. These features are utilized to characterize faces. Finally, in the hybrid method, both the holistic and local methods are utilized to perceive and distinguish a face [3]. The essential challenge confronting any calculation for facial recognition is the deficit in the appearance of the face that is expected to perceive. The motivation behind this work is to construct complete face images by utilizing the stitching image algorithm.

## II. Related Work

Over the last two decades, numerous analysts have proposed and executed different display image stitching frameworks. In the authors have introduced new systems on image stitching based on the histogram-matching algorithm [4]. Histogram coordinating is utilized for image adjustment, so the images stitched have a similar level of brightness. At this point, the paper embraces the SIFT algorithm to separate the key features of the images and plays out the harsh coordinating procedure. This work followed by the RANSAC algorithm for fine matches finally ascertained the most suitable scientific mapping model between the two images and as indicated by the mapping relationship, a straightforward weighted normal algorithm was utilized for image blending.

Authors in [5] provided a specialized examination for the fast image stitching calculation based on SIFT. Firstly, the images are separated into squares. The component sorts these neighborhood image squares and is resolved. The element purposes of the nearby image squares are removed utilizing diverse streamlined techniques adaptively. Secondly, we utilize coordination to achieve the changed framework and the RANSAC algorithm connected to expel incorrect coordinating point sets. Finally, the stitched image can be achieved by image blending.

Another proposal [6] has provided a new approach for image stitching techniques using (DTW). This work proposes a novel technique that uses the Dynamic Time Warping (DTW) algorithm to coordinate sets of images for image stitching. They additionally perform a measurement-reducing plan that shrinks the computational multifaceted nature of the standard DTW algorithm without influencing its execution. The viability of their proposed technique is shown in the stitching of 50 sets of restorative X-beam images and its execution contrasted with those of standardized cross-relationship (NCC), minimum average correlation energy (MACE) channels, total of-square-contrasts (SSD), and the entirety of absolute-contrasts (SAD). Their technique likewise beats two generally utilized stitching programs accessible on the web called Hugin and Auto-stitch.

The work in [7] gives an in-depth review of the current image mosaicing algorithms by ordering them into a few classes. For each class, the principal ideas will be clarified, and afterward, the adjustments made to the fundamental ideas by various analysts clarified. Moreover, this paper additionally discusses the focal points and burdens of all the mosaicing classes. Several previous investigations have led the field of face detection and recognition in settled images through a different scheme of frameworks. Shahin and EL-SAYED proposed a face recognition system based on the Geometrical Approach. The geometrical approach consists of two phases, namely, the detection phase and the recognition phase. The edge of the tested face is detected in the first phase after some pre-processing operations have occurred on the tested image. The second phase attempts to identify the angles of the face outline. After calculating the angle vector of the tested face, it will compare with the training set angles vectors via a neural network to identify the face [8].

Recognizing the weaknesses and strengths of machine learning techniques is critical for real life applications as well as being a prerequisite for determining extensive research and development requirements. Papers on intense analysis model can be found in literature as part of either job that specifically focuses on the attributes of deep models, or ii) work which explores the attributes of deep models as component of that other participation. Papers in the first collective, like ours, typically investigate different models and current legal findings that address a variety of deep models as their main contribution, whereas papers with in second category introduce a new classification model and then analyze its attributes.

This work proposed a discriminative feature-learning approach to recognize the face using convolution neural networks (CNNs) [9]. The proposed center loss algorithm is needed for the task of face recognition. In this algorithm, they train robust CNNs to obtain the deep features with two key learning objectives, intra-class compactness, and inter-class dispensation, as much as possible, which are necessary to the process of face recognition. Kasar et al [10] proposed a strategy for face recognition using artificial neural networks (ANN). They examined the face recognition methods proposed by numerous specialists utilizing ANN, which are used as a part of the field of pattern recognition and image processing.

Furthermore in [11] proposed a strategy for face detection and recognition based on (PCA) – (LDA) and square Euclidean distance with the Viola-Jones algorithm. Their proposed strategy is based on the appearance-construct features that concentrate on the whole face image as opposed to neighborhood facial features. The initial phase in the face recognition framework is face location. The Viola-Jones' face location technique equipped for handling images to a great degree while accomplishing high identification rates is utilized. Feature extraction and measurement-reducing strategy connected after face recognition. The principal component analysis (PCA) technique is generally utilized as a part of pattern recognition. The Linear Discriminant Analysis (LDA) technique, used to overcome the disadvantages of PCA, has effectively connected to face recognition. It is accomplished by projecting the image onto the Eigenface space by PCA and then applying unadulterated LDA over it. Subsequently, Square Euclidean Distance (SED) was utilized. This distance classifier is required to identify the similarity between the tested face images with those located in the training set.

Finally, conducted a comprehensive survey on pose-invariant face recognition [12] (PIFR). They discussed the intrinsic challenges in PIFR and exhibited a complete audit of built-up systems. They characterized the current PIFR strategies into four classes, namely, pose-robust feature extraction methodologies. They described and assessed the inspirations, systems, geniuses/cons, and the execution of agent approaches.

The key contributions of the proposed work are summarized as,

- Initially, the face images dataset is trained in a system.
- At first, an input testing image is taken from the dataset.

- Consequently, image stitching process operation is performed to determine the missing part of the face.
- Moreover, feature extraction is done through the Eigen face approach through PCA and geometrical method.
- Finally, the displacement classifier that utilizes Square Euclidean and City-block distances is used for categorizing the faces.

### III. METHODOLOGY

The proposed system developed in this paper is divided into two phases. The first phase is stitching the face image for the face that needs to recognize, while the second phase is face recognition. Fig. 1 illustrates the scheme of the proposed system.

Initially, the input face image was taken from the dataset. The image undergoes a face-stitching operation to find the missing part of the face. Through this, the complete face image is obtained, as the human faces are symmetrical on either side. After the image stitching operation, the features extraction is accomplished to classify the face image. Using the selected features, the face image is categorized as either known or unknown face by a distance classifier, which utilizes the Eigen-face approach with PCA and geometrical method.

#### A. Image Stitching

Image stitching is the method utilized for obtaining a more extensive field of perspective of a scene from a succession of halfway perspectives. It is an alluring exploration zone given its extensive variety of applications, including movement identification, determination upgrade, checking worldwide land use, and face insertion. These procedures can be classified into two groups: direct techniques and feature-based systems as shown in Fig. 2.

In a direct technique, every pixel located in the image is compared with each other, which is an exceptionally complex method. The direct way to perform an alignment between two images is to shift one image that corresponds to another by comparing pixels of the two images under testing. This comparison will depend on the rows or columns of each image and the mean square error (MSE) for each row or column will be calculated and will take as a reference to compare the given images. However, MSE is an example of the error metric. The [13] principal obstacle of direct methods is that they have a constrained scope of union.

The direct method utilizes data from all pixels. It iteratively refreshes a gauge of homography with the goal that a specific cost of work is limited. To speed up the error metric search process, hierarchical motion estimation is used. In hierarchical motion, an image pyramid is first [14] created and a search process over a fewer number of discrete pixels will perform at coarser levels.

Thirdly, Fourier-based [15] stitching depended on performing a convolution in the spatial domain resembles the summation of one signal with its conjugating of the other. In a parametric motion, a single constant translation vector with a correspondence map will be used. Finally, due to the feature points that may not be accurately located, an incremental motion refinement algorithm can calculate a more accurate matching score. However, incremental motion refinement needs more calculations than other algorithms [16], so they consider time-consuming, which reflects a decrease in its performance.

In contrast, feature-based systems are progressively more prominent and broader in mosaicing. This is particularly the result of the quality of new calculations and types of invariant features that have evolved over recent years. In Schmid represent a survey on key points detection and implements an experimental comparison to determine the repetitive features of detectors [17] and the information content available at each detected key point. The feature matching process will occur after detecting the features and key points. In addition [18] the feature matching process will determine which feature comes from locations in different images. The fundamental qualities of strong locators incorporate invariance to image scale invariance, interpretation invariance, and turn changes [19].

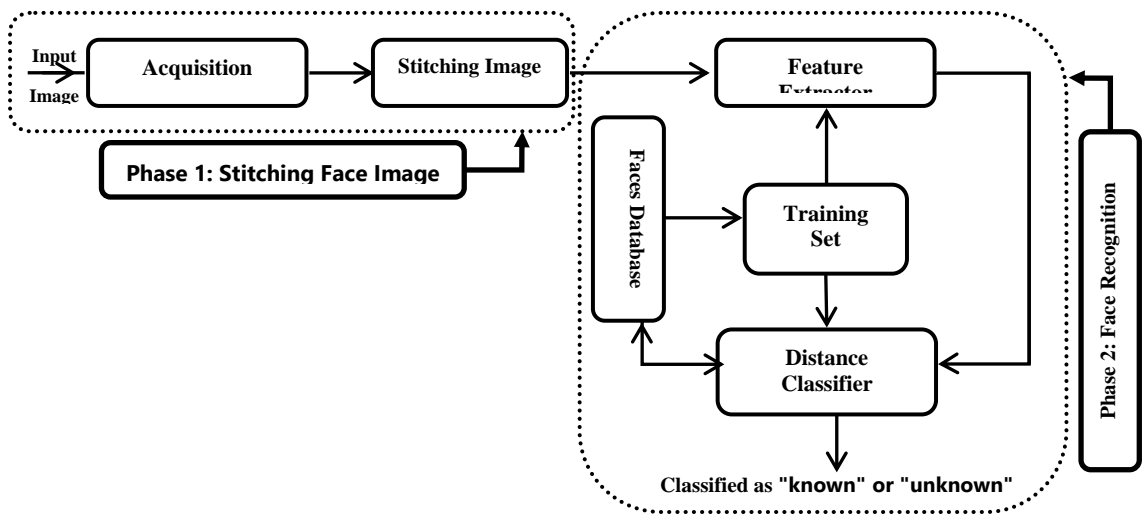

Fig. 1. Outline of the Typical Face Recognition System.

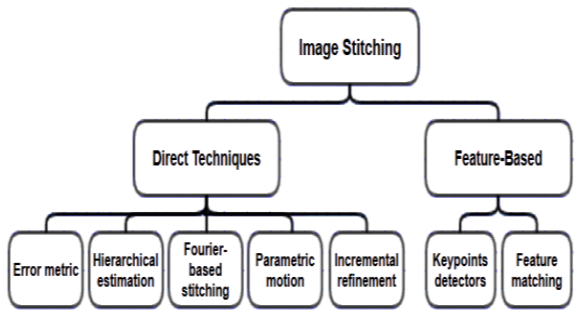

Fig. 2. Images Stitching Methodology.

Phase 1: Stitching the Face Image.

In this phase, the process of stitching the face image will be accomplished and completed according to the steps summarized in Table I.

TABLE I. PROPOSED FACE STITCHING APPROACH

1: Tested Image Acquisition: the image of the face to identify is processed. The facial image in this work assumed an incomplete face, which represents the main challenge in any face recognition system.

2: Nose Area Detection: the human face is symmetric around a certain line. To find such a line, the process of the nose detector is required. Here, the Viola-Jones Nose Detection Algorithm [20] was used. Then the centroid of the nose region is equivalent to the centroid of the rectangle that surrounds it. The centroid lies at the intersection point of the rectangle diagonals. The diagonals intersect at height $(\frac{h}{2})$ from reference y-axis and at width $(\frac{b}{2})$ from reference x-axis.

3: Vertical Line Drawn: A vertical line is drawn perpendicular to rectangle width (b) passing through the centroid. The whole face is symmetric around this vertical line.

4: Cropped Image: The image of the uncompleted tested face image cropped to obtain a cropped image $I_1$. The dimension of the cropped image $I_1$ from a pixel that has coordinate x = 0 to width equals $(x_0 + \frac{b}{2})$ from reference x-axis and from a pixel that has coordinate y = 0 to height equals the height of the vertical line that drawn in the previous step, which is equal to the height of the original tested image from reference y-axis. Whereas $(x_0, y_0)$ represent the origin, i.e., [21] coordinate of the first pixel in the cropped nose detector image.

5: Flipped Image: A horizontal reflection of $I_1$ is performed to generate the missing part of the face, this part will be denoted as $I_2$.

6: Stitched Image: The first step is to calculate the relative positions of the obtained images and to produce a vacant set of images in the computer memory where these images will be assigned. The following stage is identifying the purpose of best correlation, which is performed by sliding contiguous image edges in the two headings until the point where the best match of edge features is found. The normalized cross-correlation coefficient for the case above is defined as in equation (1):

$$\text{Cross Correlation} = \frac{\sum_{x=0}^{L-1}\sum_{y=0}^{K-1}(w(x,y)-\bar{w})\left(f(x+i,y+j)-\bar{f}(i,j)\right)}{\sqrt{\sum_{x=0}^{L-1}\sum_{y=0}^{K-1}(w(x,y)-\bar{w})^2}\sqrt{\sum_{x=0}^{L-1}\sum_{y=0}^{K-1}\left(\left(f(x+i,y+j)-\bar{f}(i,j)\right)\right)^2}} \quad (1)$$

Where $w(x, y)$ represents a pixel value of the image to place; $\bar{w}$ is the mean value of all pixels included in the selected – cropped -box area $f(x + i, y + j)$ represents a pixel value of the composite image inside the box area. However, $\bar{f}(i, j)$ are the mean value of all pixels of the composite image within the box area and parameters K, L represents the box dimensions in the number of pixels included [21].

7: Image Blending: After all the input images had aligned with each other, we will use a multi-band image blending approach to produce seamless panoramic views by choosing a suitable compositing surface.

Fig. 3 depicted the outline of the image stitching process. Fig. 3(a) depicts the input test image. To find the symmetric line of the face, the nose detection technique is used and the nose-detected image is shown in Fig. 3(b). In addition, the centroid point of the nose is shown in Fig. 3(c) and the vertical line through that point is given in Fig. 3(d). The image is cropped through that vertical line to obtain the cropped image and it is shown in Fig. 3(e). Now, the image flipped to obtain the reflected part of the cropped image. It is shown in Fig. 3(f). The search area of the stitch is shown in Fig. 3(g). After face stitching is done, the complete face image would be obtained and it is shown in Fig. 3(h) and Fig. 3(i) is the final face image after the image blending operation is performed.

*B. Face Recognition Techniques*

Face Recognition has been an interesting topic for many computers science engineers who deal with artificial intelligence. The computer first detects the face and then for recognition, a step will be performed. Face recognition is considered a pattern recognition task performed precisely on human faces [8]. The outcome of this process is to classify either a face as "known" or "unknown" which compares the given unknown faces with stored known faces. The face image must be with a uniform background to avoid problems concerning the background complexities. However, it may be affected by the change in facial expressions. For example, laughter and crying change the mouth and eyes opening size, and aging also plays an important role as the face detail changes. Much research in computer recognition of faces has focused on identifying individual face features such as the nose, eyes, head outline, and mouth. The closest match between stored data and face image achieves recognition.

PCA is a standard technique used to distinguish patterns and signal processing. It is a statistical method employed to reduce data dimensions and extract features, which is an essential step in facial recognition. The analysis of basic compounds involves a mathematical procedure that converts several interrelated variables into several non-interrelated variables called basic components. These components are linked to the original variables by orthogonal transformation and are defined in such a way that the first primary component has the highest variance and the second fundamental component has the second-highest contrast and so on. Go back to Osama Shahin [8] present a face recognition approaches that depend on the geometry of the head which is done by calculating the angles for the head circumference and then storing these readings in a vector that describe a given face and will be used for comparison with other vectors that represents other cases.

Phase 2: Face Recognition Algorithm

The idea of the proposed system is to identify human faces if they are recorded in the database of the system as well as categorize individuals whose images are not recorded in the database as unqualified or as strangers through the process of automatic identification [22] and identification of persons. In this phase, Eigenfaces for recognition algorithm and Geometrical Approach for Face Detection and Recognition [8] will be used.

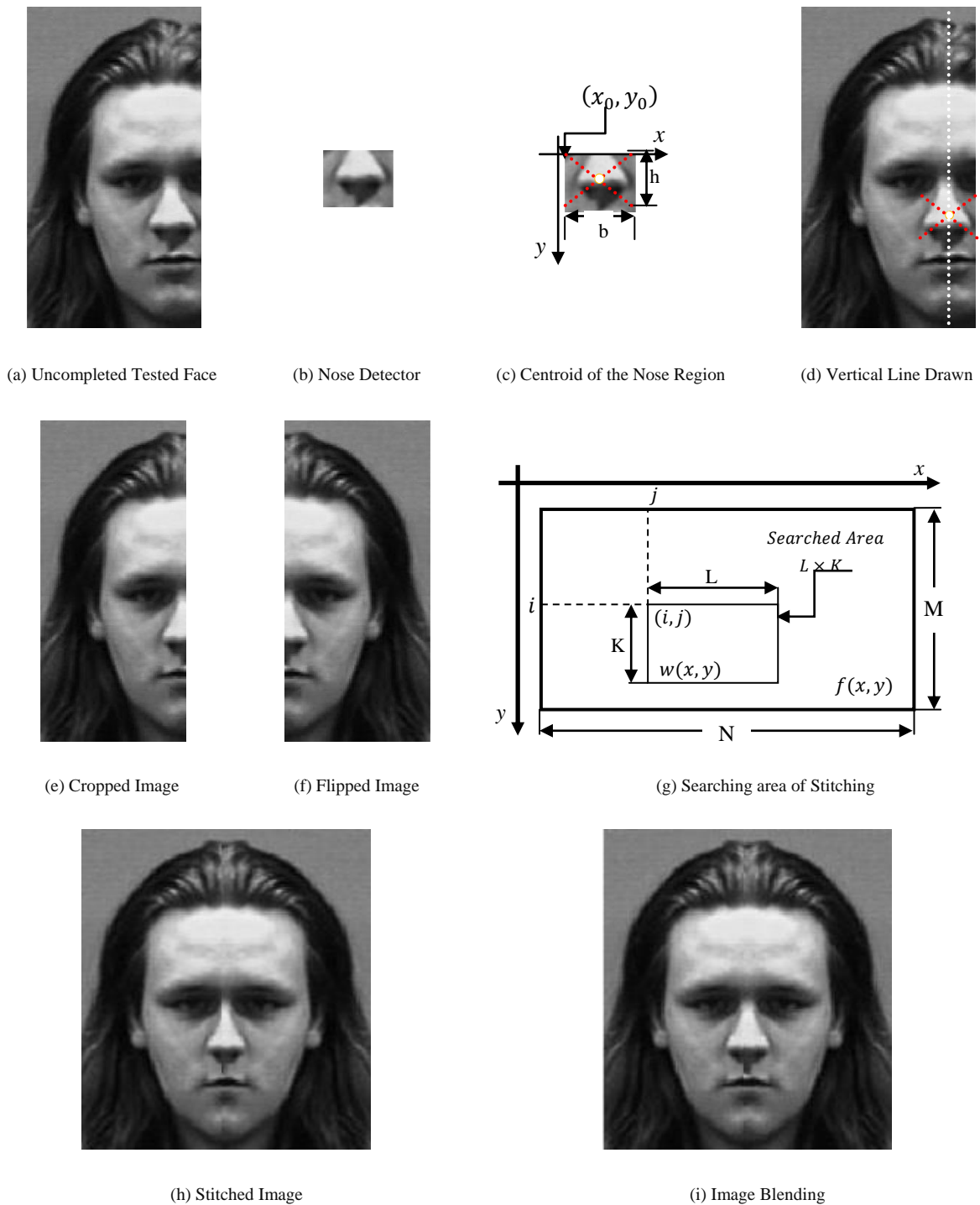

Fig. 3. Outline of the Image Stitching Process.

## IV. Experimental Results

The system was tested on database FACES94. The training dataset contains a total number of 3080 images of 123 individuals grouped and classified into three categories (male, female, and male staff) taken with a little variation in head position. The image had a resolution of 180×200 datasets [23]. The training sample of face images is shown in Fig. 4(a). These training sample images normalized to minimize blunders caused by lighting conditions. The normalized face images as shown in Fig. 4(b).

Fig. 5 shows the test image that is fed into the proposed system for classification. The quality of the image stitching process is located by observing the seam lines between the stitched images.

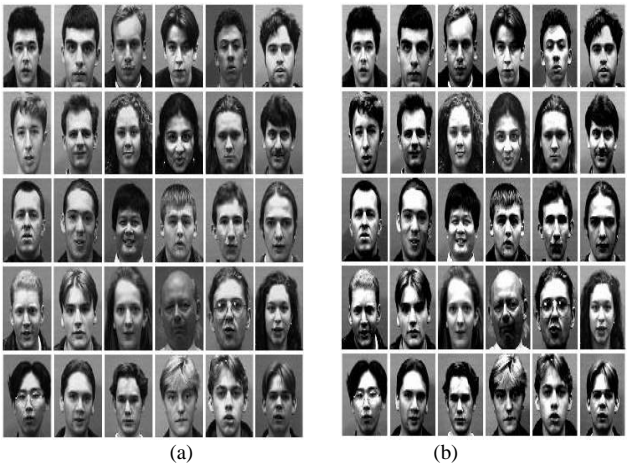

(a)        (b)

Fig. 4. (a): Initial Training, (b): Normalized Training Image Set.

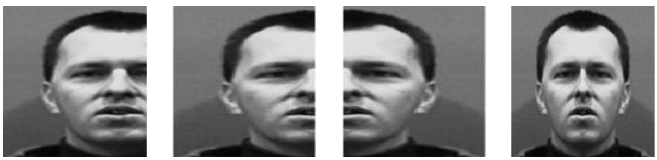

(a) Tested Image   (b) Left Half   (c) Flipped Image   (d) Stitched Image

Fig. 5. Tested and Stitched Image.

As previously mentioned, the experiments were performed on the FACES94 database with various numbers of training images. The percentage of discrimination calculated using the method of analyzing only the basic compounds in the extraction of properties and neural networks in discrimination and depending on the mainframe. The percentage of the recognition rate [24] is dependent on the number of Eigenfaces with the highest value of Eigenvalues. When 100 Eigenfaces were taken, the recognition rate was 94.2% with the Euclidean distance classifier and was equal to 92.3% when the City-Block classifier is used. When taking 150 Eigenfaces, the recognition rate rose to 94.6% with the Euclidean distance classifier and 93.1% with the City- Block classifier. This percentage then dropped to 92.2% and 91.8% respectively and then to 91.7% and 90.4% with the Euclidean and City-Block classifiers respectively when 200 and 250 Eigenfaces were taken. On taking 300 Eigenfaces, the percentage increased again to 95.1% and 93.7% respectively when Euclidean and City-Block classifiers were used. Fig. 6 and Table II show the result of the recognition process using Eigenfaces.

A comparative recognition error rate for glasses-persons is wearing glasses – and no glasses – persons are not wearing glasses – recognition using the eigenfaces and geometrical approaches methods are depicted in Table III.

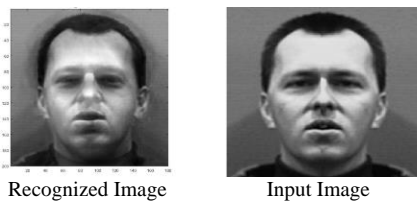

Recognized Image      Input Image

Fig. 6. Correct Face Recognition Result for FACES94 Database.

TABLE II. RECOGNITION RATES FOR THE PROPOSED ALGORITHM

| No. of Eigenfaces | Recognition Rate % Euclidean Distance | Recognition Rate % City-Block Distance |
|---|---|---|
| 100 | 94.2 | 92.3 |
| 150 | 94.6 | 93.1 |
| 200 | 92.2 | 91.8 |
| 250 | 91.7 | 90.4 |
| 300 | 95.1 | 93.7 |

TABLE III. A COMPARATIVE RECOGNITION ERROR RATES

| Approach | Error rate % for glasses | Error rate % for no glasses |
|---|---|---|
| Eigenfaces approach | 45.2 | 20.2 |
| Geometrical approach | 25.5 | 7.7 |

After the process of the recognition is performed, the calculation of Image Quality Metrics (IQM) between the stitched image and the original one is needed. In this work, MSE and CR [25] were chosen as quality factors. Table IV shows the advantages and disadvantages of several existing face detection techniques.

TABLE IV. ADVANTAGES AND DISADVANTAGES OF VARIOUS EXISTING FACE DETECTION TECHNIQUES

| Methods | Advantages | Disadvantages |
|---|---|---|
| Feature face detection | More precise and low execution time | Maximum learning time |
| Geometric face detection | Effective approach and easy to implementation | Low precise and more false alarm |
| Haar like feature face detection | Feature extraction has been improved, and there are very few false alarms now. | Maximum execution time and implementation difficulty |

### A. Mean Square Error (MSE)

MSE is needed to measure image quality. Mean Square Error must be zero in a perfect case because it is the difference between the original images and the stitched ones. However, in a practical case the smaller the value of MSE, the better the quality of the stitched image.

$$\text{MSE} = \sum_{i=1}^{m} \sum_{j=1}^{n} \frac{OI(i,j) - SI(i,j)}{m*n} \quad (2)$$

Where "OI" is the original image, "SI" is the stitched image, "M" & "N" is the numbers of rows and columns in both images respectively.

### B. Correlation Coefficient (CR)

CR was used in measuring closeness between the original image and the stitched one. The correlation Coefficient should equal 1 in a perfect case. However, in a practical case value of CR near one is a significant figure.

$$C_r = \frac{\sum_m \sum_n (X_i - \bar{X})(Y_i - \bar{Y})}{\sqrt{(\sum_m \sum_n (X_i - \bar{X})^2)(\sum_m \sum_n (Y_i - \bar{Y})^2)}} \quad (3)$$

Where "$\bar{X}$" & "$\bar{Y}$" is the original image & stitched image average values respectively, "M" & "N" are the numbers of rows and columns in both images respectively. Table V depicted MSE and Cr, the calculations for a sample of stitched and original images.

The accuracy and error rate comparison of the existing and proposed mechanism is shown in Fig. 7(a) and (b).

From the above figure, it is clear that the proposed method achieves higher accuracy of 99.78% and a reduced error rate of 0.22% compared to the existing CNN [9] and Multibatch estimator method approaches [26].

TABLE V. MSE AND CR ARE CALCULATIONS FOR A SAMPLE OF STITCHED AND ORIGINAL IMAGES

| Cases | | MSE | Cr |
|---|---|---|---|
| Stitched Image | Original Image | | |
| 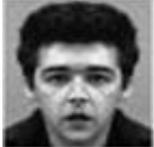 | 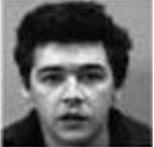 | 7.2011 | 0.9902 |
| 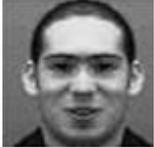 | 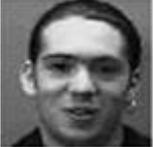 | 13.2465 | 0.9734 |
| 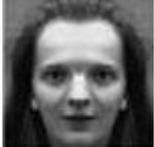 | 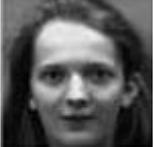 | 12.3301 | 0.9855 |
| 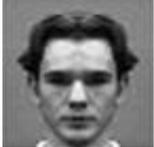 | 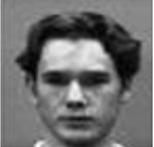 | 9.3312 | 0.9977 |
| 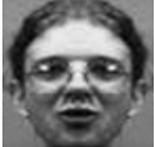 | 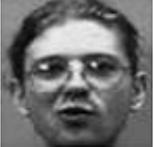 | 8.2135 | 0.9924 |

The proposed image stitching approach with Eigen-face and geometrical approach is compared with the existing CNN [9], Voting, and Random Subspace with Random Forest approaches [26]. The comparison of proposed and existing face recognition mechanisms is framed in Table VI.

TABLE VI. COMPARISON OF PROPOSED AND EXISTING FACE RECOGNITION METHODS

| Parameter | Accuracy | Error rate |
|---|---|---|
| CNN | 99.28 % | 0.72 % |
| Multibatch estimator method | 98.20 % | 1.80 % |
| Proposed | 99.78 % | 0.22% |

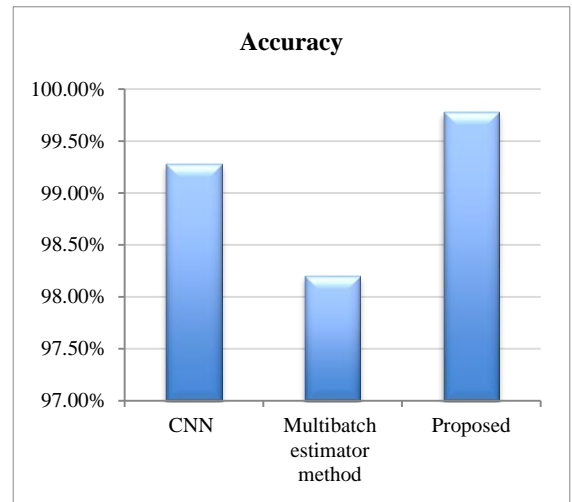

(a)

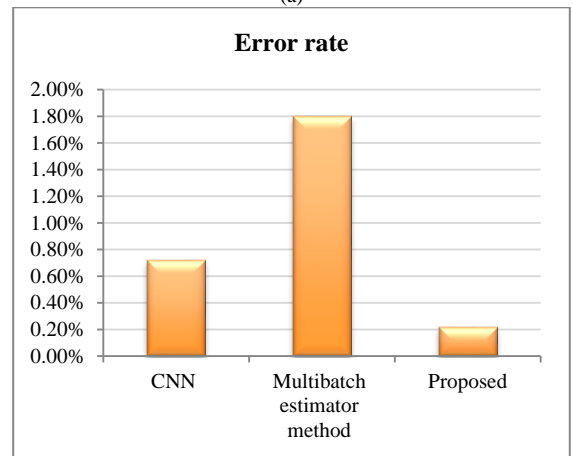

(b)

Fig. 7. (a) Accuracy Comparison of Existing and Proposed Methods, (b) Error rate Comparison of Existing and Proposed Methods.

## V. CONCLUSION

This work proposes an improved algorithm for human face recognition by using two approaches: Eigenfaces and geometrical methods from part of a facial image, which is based on image stitching. The geometrical approach was accurate especially when the person wears glasses than Eigenface, but the Eigenface approach is quick and simple for the face recognition problem. Face recognition from the perspective of PCA connected on two distance classifier systems with a certain training dataset. The training dataset contained a total number of 3080 images of 123 individuals are grouped and classified into three categories (male, female, and male staff) and was taken with a little variation in head position. The test results identified that PCA gave the best results with the squared Euclidean distance methodology, with a recognition rate of 95.1%, which was more noteworthy than the city-Block distance strategy. The recognition rate also varied with the number of Eigenfaces used in the experiment. In future development, much improved methods can be ready for improved performance. With a larger data set, a greater number of pixels could be learned. Components that are less expensive can be suggested.

REFERENCES

[1] Ghosh and N. Kaabouch, "A survey on image mosaicing techniques," Journal of Visual Communication and Image Representation, vol. 34, pp. 1-11, 2016.

[2] S. Mistryand A. Patel, "Image stitching using harris feature detection," International Research Journal of Engineering and Technology (IRJET), vol. 3, no. 4, 2016.

[3] J. K. Essel, "Head tilt classification using FFT-PCA/SVM algorithm," PhD diss., University of Ghana, 2018.

[4] J. Zhang, G. Chen and Z. Jia, "An image stitching algorithm based on histogram matching and SIFT algorithm," International Journal of Pattern Recognition and Artificial Intelligence, vol. 31, no. 04, 2017.

[5] Chen Yue, Yan Zhao, and S. G. Wang, "Fast image stitching method based on SIFT with adaptive local image feature," Chinese Optics, vol. 9, no. 4, pp. 415-422, 2017.

[6] S. Adwana, I. Alsaleh and R. Majed, "A new approach for image stitching technique using dynamic time warping (DTW) algorithm towards scoliosis X-ray diagnosis," Measurement, vol. 84, pp. 32-46, 2016.

[7] D. Vaghela and P. Naina, "A review of image mosaicing techniques," International Journal of Advance Research in Computer Science and Management Studies, vol. 2, no. 3, 2014.

[8] O. R. Shahin and E. Ayman, "Geometrical approach for face detection and recognition," Minufiya Journal of Electronic Engineering Research (MJEER), vol. 18, no.1, 2008.

[9] Y. Wen, K. Zhang, Z. Li and Y. Qiao, "A discriminative feature learning approach for deep face recognition", In European Conference on Computer Vision Springer, pp. 499-515, 2016.

[10] M. M. Kasar, D. Bhattacharyya and T. H. Kim, "Face recognition using neural network: a review," International Journal of Security and Its Applications, vol. 10, no. 3, pp. 81-100, 2016.

[11] N. H. Barnouti, S. S. Al-Dabbagh, W. E. Matti and M. A. Naser, "Face detection and recognition using viola-jones with PCA-LDA and square euclidean distance," International Journal of Advanced Computer Science and Applications (IJACSA), vol. 7, no. 5, pp. 371-377, 2016.

[12] C. Ding and D. Tao, "A comprehensive survey on pose-invariant face recognition," ACM Transactions on Intelligent Systems and Technology (TIST), vol. 7, no. 3, 2016.

[13] S. Pravenaa and R. Menaka, "A methodical review on image stitching and video stitching techniques," International Journal of Applied Engineering Research, vol. 11, no. 5, pp. 3442-3448, 2016.

[14] C. Guo, L. Wang and F. Deng, "The auxiliary model based hierarchical estimation algorithms for bilinear stochastic systems with colored noises," International Journal of Control, Automation and Systems, vol.18, no. 3, 2020.

[15] R. Perrot, P. Bourdon and David Helbert, "Confidence-based dynamic optimization model for biomedical image mosaicking," JOSA, vol. 36, no. 11, 2019.

[16] R. Szeliski, "Feature detection and matching in computer vision," Texts in Computer Science. Springer, London, 2011.

[17] Ma, J., X.Jiang, A.Fan, J.Jiang, and J.Yan, "Image matching from handcrafted to deep features," International Journal of Computer Vision, 2020.

[18] M. Karpushin, G. Valenzise and F. Dufaux, "Good features to track for RGBD images", IEEE International Conference on Acoustics, Speech and Signal Processing (ICASSP), pp. 1832-1836, 2017.

[19] M. Z. Bonny and M. S. Uddin, "Feature-based image stitching algorithms", International Workshop on Computational Intelligence (IWCI), pp. 198-203, 2016.

[20] K. Vikram and S. Padmavathi, "Facial parts detection using viola jones algorithm," in Proceedings of the Advanced Computing and Communication Systems (ICACCS), pp. 1-4,2017.

[21] V. Rankov, R. J. Locke, R. J. Edens, P. R. Barberand B. Vojnovic, "An algorithm for image stitching and blending," In Three-Dimensional and Multidimensional Microscopy: Image Acquisition and Processing XII, vol. 5701, pp. 190-199, 2005.

[22] A. L. Machidon, O. M. Machidon and P. L. Ogrutan, "Face recognition using eigenfaces, geometrical PCA approximation and neural networks," in Proceedings of the 42nd International Conference on Telecommunications and Signal Processing (TSP), pp. 80-83, 2019.

[23] The Database of FACES94, http://cmp.felk.cvut.ci/space lib/faces/faces94.html.

[24] O. R. Shahin and M. Alruily,"Vehicle identification using eigenvehicles," in Proceedings of the IEEE International Conference on Electrical, Computer and Communication Technologies (ICECCT), pp. 1-6, 2019.

[25] U. Sara, M. Akter andM. S. Uddin, "Image quality assessment through FSIM, SSIM, MSE and PSNR—a comparative study," Journal of Computer and Communications, vol. 7, no. 3, pp. 8-18, 2019.

[26] O. Tadmor, Y. Wexler, T. Rosenwein, S. Shalev-Shwartz and A. Shashua, "Learning a metric embedding for face recognition using the multibatch method", in Advances in Neural Information Processing Systems, pp. 1388–1389, 2016.